*Review*

# Corpus-Based Paraphrase Detection Experiments and Review

Tedo Vrbanec [1,*] and Ana Meštrović [2,3,*]

1 Faculty of Teacher Education, University of Zagreb, Savska Cesta 77, 10000 Zagreb, Croatia
2 Department of Informatics, University of Rijeka, Radmile Matejčić 2, 51000 Rijeka, Croatia
3 Center for Artificial Intelligence and Cybersecurity, University of Rijeka, Radmile Matejčić 2, 51000 Rijeka, Croatia
* Correspondence: tedo.vrbanec@ufzg.unizg.hr (T.V.); amestrovic@inf.uniri.hr (A.M.)

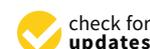

**Abstract:** Paraphrase detection is important for a number of applications, including plagiarism detection, authorship attribution, question answering, text summarization, text mining in general, etc. In this paper, we give a performance overview of various types of corpus-based models, especially deep learning (DL) models, with the task of paraphrase detection. We report the results of eight models (LSI, TF-IDF, Word2Vec, Doc2Vec, GloVe, FastText, ELMO, and USE) evaluated on three different public available corpora: Microsoft Research Paraphrase Corpus, Clough and Stevenson and Webis Crowd Paraphrase Corpus 2011. Through a great number of experiments, we decided on the most appropriate approaches for text pre-processing: hyper-parameters, sub-model selection—where they exist (e.g., Skipgram vs. CBOW), distance measures, and semantic similarity/paraphrase detection threshold. Our findings and those of other researchers who have used deep learning models show that DL models are very competitive with traditional state-of-the-art approaches and have potential that should be further developed.

## 1. Introduction

Paraphrasing is the process of rewriting text to change the form and expression while retaining its original meaning. Automatic paraphrase detection has an important role in the various tasks, including plagiarism detection, authorship attribution, question answering, text summarization, text mining in general, etc. Likewise, the somewhat more general task of the measuring of the semantic similarity of texts is significant in the domain of natural language processing (NLP).

Some of the existing paraphrase systems have performed quite well; however, there are certain challenges with paraphrase detection. For example, existing paraphrase systems deliver relatively good results for clean texts, but they do not perform well when applied to noisy texts [1–3]. Moreover, in recent years there was an expansion of deep neural network models' application to the NLP domain, and that opens up a complete new field for experimentation and improvement of the existing approaches. These are deep learning (DL) models whose vectors, presenting words or documents, implicitly contain some semantic meaning. These models have become possible and usable because of advances in algorithms and computer hardware. Some of the most popular DL models are Word2Vec [4], Doc2Vec [5], GloVe [6], and FastText [7]. Business giants such as Facebook and Google quickly became involved and after positive experimental results; they began using and developing new DL models—using them in their large-scale applications, such as text translation, text analysis, facial recognition, targeted advertising, and multiple AI applications.

The main goal of this research was to find which of the DL models—from the set of corpus-based approaches—is the best solution for the task of paraphrasing detection. The motivation behind this study is to develop a framework for plagiarism detection based on statistics and machine learning, especially deep learning.

The main contributions of this paper are (i) the systematic performance overview of corpus-based models and (ii) the report of the best results/parameters/thresholds based on experimenting with numerous variations of text pre-processing techniques and hyper parameters.

In this research we compare eight models; namely, LSI, TF-IDF, Word2Vec, Doc2Vec, GloVe, FastText, ELMO [8], and USE [9]. We evaluated these models in terms of accuracy, precision, recall, and F1 measure on three different public available corpora: Microsoft Research Paraphrase Corpus (MSRP), Clough and Stevenson (C&S), and Webis Crowd Paraphrase Corpus 2011 (Webis-CPC-11). We tested various approaches by varying various text pre-processing scenarios, hyper-parameters, sub-model selection (e.g., Skipgram vs. CBOW), distance measures, and semantic similarity/paraphrase detection threshold.

The article is organized as follows: Section 2 presents related work on mostly corpus-based semantic similarity measures; Section 3 describes the path from word similarity measures to paraphrase detection—as one manifestation of the semantic similarity of texts—and then presents the methodology used in our experiments including the models and evaluation measures. The same section discusses the experimental setup, including the software environment, corpora metadata, and pre-processing text from the corpora. Section 4 presents the results of the previously explained approach, and a discussion of the results obtained. Section 5 draws conclusions and outlines future research. Afterward, the Appendix A contains precision/recall diagrams and receiver operating characteristic curves, while the second and the third contain empirical evidence of some of the claims in the article.

## 2. Related Work

There are many approaches to paraphrase detection, but most are corpus or knowledge-based or a combination of both. In our research, we have selected and described corpus approaches and a combination of corpus and knowledge-based approaches.

The use of deep neural networks for natural language processing has increased considerably over recent years. Most of the previous work on sentence modelling focused on features like n-gram overlap features [10], syntax features [6,11], and machine translation based features [10]. Recently, deep learning-based methods have shifted researchers' attention towards semantically distributed representations. A variety of deep neural network-based architectures have been proposed for sentence similarity, a strategy we also focus on in this paper. Substantial work has been carried out on paraphrase detection from the clean-text Microsoft Research Paraphrase corpus (MSRP).

Corley and Mihalcea (2005) [12] introduce a knowledge-based method that combines word-to-word semantic similarity metrics into a text-to-text metric. They used six word similarity metrics: Leacock and Chodorow, Lesk, Wu and Palmer, Resnik, Lin, and Jiang and Conrath. (WordNet-based), and the implementation of these metrics is defined between concepts, rather than words, but they used them as a word-to-word similarity metrics by selecting, for any given pair of words, the two meanings that led to the highest concept-to-concept similarity.

Mihalcea et al. (2006) [13] presented a method that outperforms a vector-based similarity approach for measuring the semantic similarity of texts using two corpus-based and six knowledge-based measures of word similarity, which they used "to derive a text-to-text similarity metric". Mihalcea also participated in the study of Corley et al. (2007) [11], wherein the researchers used "six different metrics, selected mainly for their observed performance in natural language processing applications [ . . . ] and for their relatively high computational efficiency". Those were WordNet-based word similarity metrics: Leacock and Chodorow, Lesk, Wu and Palmer, Resnik, Lin, and Jiang and Conrath. The authors combined "metrics of word-to-word similarity and language models into a formula that is a potentially good indicator of the semantic similarity of the two input texts".

Fernando and Stevenson (2008) [14] made an algorithm for paraphrase identification which makes extensive use of word similarity information derived from WordNet. For calculating similarities between pairs of sentences, they used cosine similarity of vectors presenting sentences and semantic similarity matrixes containing information about the similarities of word pairs derived from six WordNet hierarchy-based metrics. Each sentence is represented as a binary vector (with elements equal to 1 if a word is present and 0 otherwise) $\vec{a}$ and $\vec{b}$. The similarity between these sentences can be computed using the formula $sim(\vec{a}, \vec{b}) = \frac{\vec{a} W \vec{b}}{\|\vec{a}\| \|\vec{b}\|}$, where W is a semantic similarity matrix containing information about the similarity of word pairs. Formally, each element $w_{ij}$ in W represents the similarity of words $p_i$ and $p_j$ according to a lexical similarity measure.

Callison-Burch (2008) [15] used bilingual phrase pairs and altered the commonly-used phrase extraction algorithm to extract phrase labels from phrase pairs to be able to generate paraphrase or to detect them in bilingual corpora. Their major innovations are to place syntactic constraints on the common paraphrasing technique that extracts paraphrases, and that phrases are required to be the same syntactic type as the phrase that they are paraphrasing.

Chong et al. (2010) [16] applied several NLP techniques on short paragraphs to analyze the structure of the text to automatically identify plagiarized texts. They proved that NLP techniques can improve the accuracy of detection tasks, although some challenges remain, such as multilingual detection, synonymy generalization (word sense disambiguation), and sentence structure generalization.

Socher et al. (2011) [17] introduced an unsupervised feature learning algorithm based on unsupervised recursive auto-encoders which learn feature vectors for phrases in syntactic trees. These features are used to measure the word and phrase-wise similarity between two sentences.

Šarić et. al. (2012) [18] used both knowledge-based and corpus-based approaches with great success at the SemEval 2012 conference. All their knowledge-based word similarity measures are based on WordNet. In order to compute the similarity score for a pair of words, authors took the maximum similarity score over all possible pairs of concepts—WordNet synsets. The authors used the "lowest common subsumer (LCS) of concepts $c_1$ and $c_2$, which represents the lowest node in the WordNet hierarchy that is a hypernym of both $c_1$ and $c_2$". They also used the NLTK library to compute the PathLen similarity and Lin similarity measures. For corpus-based word similarity, the authors used the distributional lexical semantics models, wherein "deriving the semantic similarity between two words corresponds to comparing these distributions". They employed latent semantic analysis (LSA) over a large corpus to estimate the distributions. They computed the TF-IDF and extracted word vectors from NYT corpus and Wikipedia. The authors used many methods, measures, and models, including a support vector regression model.

Chong (2013) [19] made an extensive study trying every single method and approach to measure the similarities of texts in small and large scale corpora. The author concluded the empirical studies by showing that machine learning is an essential part of any framework for detecting plagiarism.

Mikolov et al. (2013) [4] presented a generally applicable vector off-set method for identifying linguistic regularities in continuous space word representations. They showed that word representations learned by recurrent neural network language modeling [20] are very good tools for capturing these regularities. Le and Mikolov (2014) [5] also invented the "unsupervised learning algorithm that learns vector representations for variable-length pieces of texts such as sentences and documents. The vector representations are learned to predict the surrounding words in contexts sampled from the paragraph".

Deep learning (DL) was also used by Banea et al. (2014) [21]. They experimented "with traditional knowledge-based metrics, as well as novel corpus-based measures based on deep learning paradigms, paired with varying degrees of context expansion".

Socher (2014) [22] introduced recursive deep learning methods, which are variations and extensions of unsupervised and supervised recursive neural networks (RNNs). This method uses the idea of the hierarchical structure of the text and encodes two word vectors into one vector by auto-encoder networks. Socher also presents many variations of these deep combination functions, such as recurrent neural network (RNN) and matrix-vector recursive neural networks (MV-RNN).

In [23], Kong et al. (2014) tried to detect high obfuscation plagiarism with a logical regression model. The proposed model integrated lexicon features, syntax features, semantics features, and structure features which were extracted from suspicious documents and source documents.

Gipp (2014) [24] introduced citation-based plagiarism detection, which does not consider text similarity alone, but uses citation patterns within documents to identify potentially suspicious similarity among texts. The idea for CbPD originated from the observation that plagiarists commonly disguise academic misconduct by paraphrasing copied text, but typically do not substitute or significantly rearrange the citations.

Kim et al. (2014) [25] described a series of experiments with convolutional neural networks built on top of word2vec and suggested that unsupervised pre-training of word vectors is very useful in NLP tasks.

Yin et al. (2015) [26] proposed a new deep learning architecture Bi-CNN-MI for paraphrase detection/identification (PI) by comparing sentences on multiple levels of granularity, using a convolutional neural network (CNN) and modelling interaction features at each level. These features are then the input to a logistic classifier for PI.

Gharavi et al. (2016) [27] proposed a "deep learning based method to detect plagiarism" in the Persian language. "In the proposed method, words are represented as multi-dimensional vectors, and simple aggregation methods are used to combine the word vectors for sentence representation. By comparing representations of source and suspicious sentences, pair sentences with the highest similarity are considered as the candidates for plagiarism. The decision on being a plagiarism is performed using a two level evaluation method" [27].

Thompson and Bowerman (2017) [28] proposed (corpus-based) methods for detecting the most common techniques used in paraphrasing texts (lexical substitution, insertion/deletion, and word and phrase reordering), and combined them into a paraphrase detection model.

Agarwal et al. (2017) [1] developed a robust paraphrase detection model based on deep learning techniques that is able to successfully detect paraphrasing in both noisy and clean texts. They proposed a hybrid deep neural architecture composed of a convolutional neural network (CNN) and a long short-term memory (LSTM) model, further enhanced by a word-pair similarity module.

Zhou et al. (2018) [29] proposed a deep learning model for paraphrase identification based on pairwise unit similarity features extracted from given pairs of text through a convolutional neural network (CNN) model and a semantic context correlation feature based on CNN and LSTM (long short-term memory).

Li et al. (2018) [30] made a paraphrase generator and evaluator for the paraphrases generated by their generator. To train the evaluator they used a reverse approach compared to the one used to train the generator, so we have some reservations about that evaluation. Their research is not comparable to our unsupervised approach.

Wu et al. (2018) [31] noticed that memory consumption of previous sentence encoding models grows quadratically with sentence length, and the syntactic information is neglected. Thus, they made a model that can utilize syntactic information for universal sentence encoding and filter out distant and unrelated words, and focus on modeling the interaction between semantically and syntactically important words.

Zablocki et al. (2018) [32] proposed a multimodal (text and image) context-based approach to learn word embeddings. In their experiments they showed that visual surroundings of objects and their relative localizations are very informative for building word representations.

Devlin et al. (2018) [33] from Google presented a novel technique named masked LM (MLM) which allows for bidirectional training in models. Previous efforts looked at a text sequence either from left to right or combined left-to-right and right-to-left training. Using such a technique they made the BERT model, which is used in a wide variety of NLP tasks, including question answering, natural language inference (MNLI), and others.

As Yang et al. (2019) [34] see it, "BERT neglects dependency between the masked positions and suffers from a pretrain-finetune discrepancy". To overcome this, they suggested XLNet, a generalized autoregressive pretraining method that (1) enables learning bidirectional contexts by maximizing the expected likelihood over all permutations of the factorization order and (2) overcomes the limitations of BERT thanks to its autoregressive formulation.

In order to improve paraphrase detection, El Desouki et al. (2019) [35] proposed a model that combines the text similarity approach with a deep learning approach using a skip-thought DL model to get the semantic vector of each sentence and then measure the vector similarity between the resulting semantic vectors using multiple similarity measures separately and in combination, as represented by Gomaa and Fahmy (2017) [36].

Ahmed et al. (2019) [37] proposed two variants of the tree structured LSTM model for representation of how a sentence builds semantically from the words, and the model gives a score depending on how similar two sentences are.

El Mostafa and Benabbou (2020) [3] made a comparative study of plagiarism detection. They concluded that most of studies are based on world granularity and use the word2vec method for word vector representation, which is their weak point, since they do not reflect the true meanings of the sentences. Similarly, Tenney et al. (2019) [38] found that existing models trained on language modeling are very useful for syntactic tasks but offer small improvements on semantic tasks.

In DL models so far, one word will always have a unique vector presentation, regardless of context. Although not concerned with paraphrasing, Shuang et al. (2020) [39] attempt to address the problem of polysemy, as one of the more important obstacles to better paraphrasing detection. They proposed convolution–deconvolution word embedding (CDWE), an end-to-end multi-prototype fusion embedding that fuses context-specific information and task-specific information.

## 3. Materials and Methods

### 3.1. From Semantic Similarity to Paraphrase Detection

The notion of semantic similarity is fundamental and widely understood in many NLP and related fields, including information retrieval, document classification, word sense disambiguation, machine translation, text summarization, paraphrase identification, etc. Likewise, a somewhat more general task, the measuring of the similarity of concepts, is of significant importance for other fields in which the similarity is defined differently, although the same methods are used. For example, the proposed methods can be used in the field of biotechnology for the determination of the similarity of the ontology of genes or the comparison of proteins based on their functions.

The terms that can be used instead of semantic similarity are semantic proximity or semantic distance as an opposite concept. It can be defined as the degree of taxonomic proximity between concepts [40].

Semantic similarity can be measured in terms of the numerical score (usually) from the segment [0,1] that quantifies similarity/proximity. In the existing research, various semantic similarity measures (SSMs) have been defined and many semantic similarity computational models have been proposed. Some of these approaches are discussed in this paper.

Semantic similarity can be calculated on different levels of granularity between words/concepts, between sentences/short texts, between paragraphs, between whole documents, and between texts in general. Moreover, there are various levels of comparing texts and documents: (i) comparing long texts (for example in the document classification tasks); (ii) comparing a short text with a long one (for example in the document retrieval task); and (iii) comparing short texts (for example in the paraphrase recognition tasks [17,41,42], tweets searching [1], image retrieval by captions [43], query reformulation [44,45], and automatic machine translation evaluation [46,47]). Semantic similarity on the words/concepts level is based on the hierarchy between concepts or words. It is usually defined for the taxonomy such as WordNet or for some more extensive ontology [40]. These measures assume as the input a pair of concepts or words, and return a numeric value indicating their semantic similarity. Based on word similarities it is possible to compute the similarities of texts according to variously defined equations [48]. Similarity measures on the document or text level are mainly based on the approaches developed in the NLP domain. Most of these approaches have their roots in machine learning.

Paraphrase detection can be formalized as a binary classification task: for two texts ($t_1$, $t_2$), determine whether they have the same/similar meaning. This task can be easily transferred into the task of determining semantic similarity between two texts. Since semantic similarity measures give numerical values, it is necessary to introduce a paraphrase detection threshold, which sets the value of similarity above which we would consider two text to be paraphrased.

In the case of paraphrase detection and eventually plagiarism detection, semantic similarity should be expressed on the text level as the final result. The score of semantic similarity between two texts may indicate the existence of paraphrasing. Semantic similarity and paraphrasing indication can be calculated both on the sentence and the paragraph level.

There are various approaches for the task of measuring semantic similarity. The two most prominent are: corpus-based and knowledge-based approaches [13].

Inspired by recent successes of deep neural networks (NNs) in the fields of computer vision [49], speech recognition [50], and natural language processing [51], in this paper we adopt a deep learning approach to paraphrase identification.

*3.2. Methodology*

Before we proceed with the semantic approaches, let us mention the oldest approach, i.e., the statistical one. Term-frequency-inverse document frequency (TF-IDF) developed by [52] is one of the oldest approaches to measure term (word) significance in the document from a corpus of documents. Words are given different weights than simple frequencies: TF-IDF measures of relevance or TF-IDF scores, depending on word frequency in the document (the more gives higher score of *tf*), and in corpus (the more it appears in various documents the less relevant became *idf*).

$$w_{ij} = tf_{ij} \times log \frac{N}{df_i}, \qquad (1)$$

where $tf_{ij}$ = number of occurrences of *i* in *j*, $df_i$ = number of documents containing *i*, and *N* = number of documents.

One of the first approaches in measuring semantic similarity between documents was the vector space model (VSM) originally proposed by [53] for the task of information retrieval domain. The purpose of the VSM is to represent each entity in the collection (letters in words, words in sentences, sentences in documents, documents in the corpus of documents) as a point in n-dimensional space; i.e., as a vector in VSM [54]. The closer the points in this space are, the more semantically similar they are and vice versa. For a given set of *k* documents $D = \{D_1, D_2, \ldots, D_k\}$, a document $D_i$ can be represented as a vector $D_i = (w_{i1}, w_{i2}, \ldots, w_{in})$. In the classical word-based VSM, each dimension corresponds to one term or word from the document set. Weights may be determined by using various weighting schemes; TF-IDF is usually used in the word-based VSM. The similarity between two documents $D_i$ and $D_j$ can be calculated as cosine similarity:

$$sim_{cos} = \cos \sphericalangle (D_i, D_j) = \frac{D_i \cdot D_j}{\| D_i \| \cdot \| D_j \|}, \qquad (2)$$

The main drawbacks of this model are high dimensionality, sparseness, and vocabulary problems. Therefore, there are various modifications and generalizations of this classical version of the VSM.

Another approach is the latent semantic analysis (LSA) proposed by [55]. It also exploits the vector space model, but uses a dimension reduction technique known as singular value decomposition (SVD) of the initial matrix. In this way LSA overcomes the high dimensionality and sparseness of the standard SVM model. The similarity between two documents is again calculated as a cosine similarity as in the previous equation or by using some other similarity measure.

To overcome the high dimensionality of the vector space model and to make it meaningful, i.e., to have some meaningful relations between vectors as words presentations, [56] designed a new approach by learning a distributed representation for words. Their model uses neural networks for the probability function to learn simultaneously (a) a distributed representation for each word along with (b) the probability function for word sequences.

One more recent approach is deep learning. Like the previous methods, in deep learning documents or texts can be represented as vectors by using the document to vector technique (Doc2Vec). Moreover, words are also represented as vectors by using the word to vector strategy (Word2Vec). There are variations in learning word vector representation: One uses the matrix decomposition method; for instance, LSA. Another uses context-base methods such as skip-grams, a continuous bag of words. Furthermore, there is an unsupervised algorithm that learns representation for documents or smaller samples of texts (paragraphs, sentences). At the end, vectors can be compared using cosine similarity or some other similarity measure.

One way to get word embedding is the Word2Vec model suggested by Mikolov et al. (2013) [4]. The emergence of the Word2Vec model has prompted researchers to create other vector models, such as Doc2Vec, FastText, GloVe, USE, and ELMO, which were used in these experiments, but new ones are still emerging. Thus, ELMO and USE models appeared during the experiments and were included in them. All of these models are called *2Vec models because they convert text (in the form of words, phrases, sentences, chapters, and entire documents) into vector form, creating n-dimensional vector spaces, where n is sometimes a fixed and sometimes adjustable integer. Training a neural network with texts from large unmarked corpora results in the creation of a VSM, using arbitrary parameters of which the most important are: vector space dimension, minimum word frequency, learning speed, and window size/context of observation of each word. Word2Vec consists of two sub models: CBOW (continuous bag-of-words) predicts a missing word if we present the missing word context to the model, while the skip-gram model predicts the context of a given word.

Although Word2Vec is great for vector representations of words, it is not designed to generate a single representation of n words, sentences, paragraphs, or entire documents. A possible approach to this problem is to first obtain a vector for each word, and then to calculate the vector average (or sum) for more complex structures. Unfortunately, this approach diminishes the value of teaching a Word2Vec model with certain data, because it reduces the value of the context. Le and Mikolov noticed this problem and suggested the Doc2Vec model [5]. They offered a model solution that allowed the conversion of variable word counts to fixed-size vector formats: the paragraph vector algorithm (the authors used two more expressions: feature vector and density vector). In essence, the Word2Vec model is still used here, but the paragraph ID vector was added as an extension of the CBOW model from Word2Vec. By teaching a neural network with word vectors, document vectors are also created. This model is called the distributed memory version of paragraph vector (PV-DM). The model remembers the context of the upcoming words, such as the paragraph title. Thus, word vectors present words, and document vectors present documents. Just as there are two sub models in the Word2Vec model (CBOW and Skipgram), in the Doc2Vec model in addition to PV-DM there is a sub model called distributed bag of words version of paragraph vector (PV-DBOW), and it matches a Skipgram from Word2Vec. PV-DBOW is faster and uses less memory than Word2Vec's Skipgram, since there is no need to store word vectors.

The two main types of word vector learning methods are global matrix factorization methods, such as LSA, and local context window models, such as Word2vec. The first method makes efficient use of statistics, but poorly finds analogies between words, which indicates the suboptimal structure of the vector spaces they create. Other types, such as Skipgrams from the Word2Vec model, successfully detect analogies between words, but make little use of corpus statistics because they are taught on separate local context windows, rather than on global coincidence matrix data. According to [6], the Word2vec model has revealed semantic and syntactic regularities, but the source of these regularities is not sufficiently clarified. They therefore introduced the GloVe (Global Vector) model, a logarithmic-bilinear regression model that combines the advantages of two major types of models: global matrix factorization and the local context window method. The model efficiently exploits statistics by learning exclusively on non-zero word-word co-occurrence matrix elements, not on the entire sparse matrix or on large corpus context windows, and creates a vector space that has meaningful substructure. The goal of training GloVe models is to obtain such vectors that their scalar product is equal to the logarithm of the word probability from the collocation matrix. The difference between the vectors in the vector space is equal to the difference in logarithms of the probability of occurrence of words in collocation. Given that the odds ratios are meaningful, this information is encoded into vector differences as differences by the logarithms of probability, i.e., the logarithm of the odds ratio.

One study [57] observed that vector representations neglected word morphology, assigning different forms of the same word a new vector, which is a limitation, especially for inflectional and morphologically rich languages characterized by large dictionaries and many rare words. Their FastText model, like Word2Vec, uses a continuous display of words, taught on large unlabeled corpora, but with the difference that each word is represented here as a bag of n-gram characters. A vector representation is associated with each n-gram character, and the words are represented by the sum of the n-gram character vector representations.

Universal sentence encoder (USE) and embeddings from language models (ELMO) are brand new deep learning approaches we used. USE was presented by [9] as a model trained with a deep averaging network (DAN) encoder that encodes variable length English text into fixed 512 dimensional embedding vectors. ELMO presented by [8] also converts input text into vector representation that is a function of the entire input sentence.

There are also other similar measures for text similarity, including explicit semantic analysis (ESA), salient semantic analysis (SSA), distributional similarity, hyperspace analogues to language, etc.

*3.3. Evaluation Framework*

We used the standard measures of accuracy, precision, recall, and *F*-measure (F1) to evaluate the results of the experiments. If we mark *TP* as true positive, *TN* as true negative, *FP* as false positive, *FN* as false negative, and N as number of all documents/texts/sentences, then:

Accuracy is the proportion of correctly classified documents (*TP* + *TN*) in a set of all documents (*N*):

$$A = \frac{TP + TN}{N}$$

Precision is the proportion of correctly classified documents (*TP*) in a set of positively classified documents (*TP* + *FP*):

$$P = \frac{TP}{TP + FP}$$

Recall is the proportion of correctly classified documents (*TP*) in a set of all positive documents (*TP* + *FN*):

$$R = \frac{TP}{TP + FN}$$

*F*-measure (F1) is the harmonic mean of precision and recall:

$$F = \frac{2PR}{P + R}$$

Our models give the result of similarity within the segment [0,1], so we defined a threshold above which we considered a pair of texts as paraphrasing, so we reduced the problem to a binary classification which could subsequently be evaluated by the F1 measure. If that were not so, we would have used the Pearson correlation.

*3.4. Experiment Setup*

In this section we will list and describe the basic factors of the experiments. The experimental setup includes the environmental setup, defining and retrieving corpora, their preparation for the experiments (separating text and descriptive information), their pre-processing (ballast text removal), using models to process such pre-processed text, the determination of the optimal cut-off value (threshold) for each model, and the generation of graphical and tabular outputs.

3.4.1. Environment

The experiment was performed using the Python programming language 2.7 (due to the longer duration of the research, finished on 3.7), on 64-bit Linux Debian 9 (finished on 10) operating system, on one laptop with an Intel i7-7th generation CPU and 8GB of RAM. We have used many optimized modules for NLP and Cython which allowed us to significantly accelerate the program at critical processing stages.

3.4.2. Corpora

All models were evaluated on three freely accessible corpora that contain short text pairs annotated as paraphrased or not-paraphrased or annotated with the degree of paraphrasing (degree of similarity).

Figure 1 shows large differences in corpora where the C&S corpus contains only 100 texts, while the MSRP and Webis contain 10,948 and 15,718 respectively. The MSRP and Webis also differ a lot from each other, as the latter, besides containing about 50 percent more text units, has on average 16.5 times more text. After pre-processing the text, the number of text units in C&S corpus remained the same, but MSRP and Webis changed considerably: the former increased because of multiple usage of the same texts while the latter decreased because it contained some empty files.

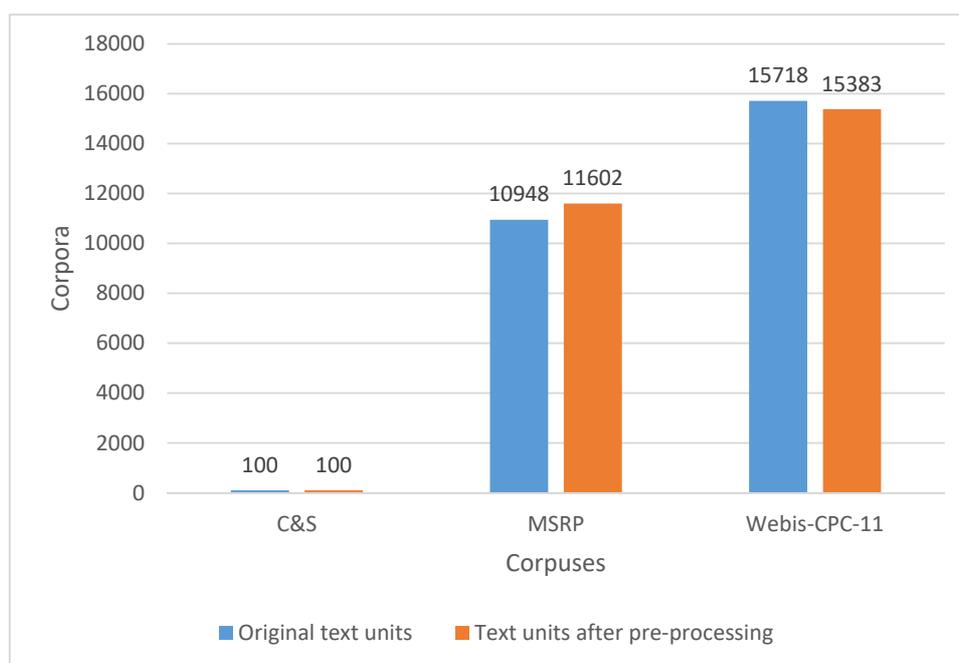

**Figure 1.** Number of text units before and after pre-processing.

Clough and Stevenson Corpus of Plagiarized Short Answers

The C&S corpus [58] provides a collection of five questions and 95 short answers on those questions. Students (respondents) were asked questions with instructions on how to answer them. The questions contained original texts taken from Wikipedia. Some of the respondents were asked to copy and paste their answers; some were asked to paraphrase the original Wikipedia texts lightly or heavily; and some were asked to use lecture notes and textbooks. Thus, the short answers include texts constructed with each of the following four strategies: (1) copying and pasting individual sentences from Wikipedia; (2) light revision of material copied from Wikipedia; (3) heavy revision of material from Wikipedia; (4) non-plagiarized answers produced without even looking at Wikipedia. Information on the affiliation of the answers to one of the categories is also available. We reduced the annotations to binary form to make the results comparable to those of other corpora.

Microsoft Paraphrase Corpus (MSRP)

This corpus consists of 5801 pairs of sentences collected over a period of 18 months from thousands of news sources on the web [59]. Each pair is accompanied by the judgment of two or three judges (the third decides if the two disagree) as to whether the two sentences are close enough to be considered paraphrases. After resolving the differences between the classifications, 3900 (67%) of the original 5801 pairs were rated semantically equivalent, so the data were unbalanced and were then split into a training set containing 4076 examples and a test set containing 1725 examples.

Webis Crowd Paraphrase Corpus 2011 (Webis-CPC-11)

The Webis contains 7859 candidate paraphrases obtained from Mechanical Turk crowdsourcing [60]. The corpus is made up of 4067 accepted paraphrases, 3792 rejected non-paraphrases, and the original texts. The original samples were extracted from Project Gutenberg, and ranged from 28 to 954 words in length.

Table 1 and Figure 2 present some descriptive statistics about corpora used in the experiment. The C&S corpus consists of 100 text files. Five of them are original texts that had to be paraphrased by instructed students in four different ways and levels. The MSRP corpus is in the shape of three files: train, test, and data. The train and test files consist of records containing five tab-separated fields: quality, #1ID, #2ID, #1String, and #2String. The data file contains records of seven tab-separated fields: sentence ID, string, author, URL, agency, date, and web date. In our experiments we used the train and test files, because they include official information whether a pair of sentences is paraphrased or not. We have split 5801 pairs of sentences into 11,602 individual documents, but some of them were used more than once, so we got 10,948 unique documents. The Webis corpus consists of 23,577 text files (3 × 7859) but we extracted and removed 1/3 (metadata file) from the table metadata as (1) the metadata can be used in more practical form and (2) the models would be negatively affected with the metadata files. Thus, every one of the 7859 paraphrase pairs is represented by three files, containing the original text, the paraphrased text, and a metadata with information about the task identifier, task author identifier, time taken, and whether the paraphrase was accepted or rejected [61]. The Webis corpus contains 335 blank files. All of these files have paraphrases in the name and the official annotation 0 indicating that the document pair is not a paraphrase.

The difference between the corpora is evident from Figure 2, which is the graphical representation of the data presented in Table 1.

The corpora had some drawbacks, such as the one regarding empty files in Webis. An even more serious remark is that the official grades from MSRP and Webis plagiarism official ratings are too often incorrect, so their usability is questionable. C&S plagiarism ratings had to be translated into 0/1 grades so the same methodology for all corpora could be used. Despite these shortcomings, we decided to use them to be able to compare them with results of previous studies.

Table 1. Corpora metadata.

| Property | C&S | MSRP | Webis-CPC-11 |
|---|---|---|---|
| Original text units | 100 | 10,948 | 15,718 |
| Text units after preprocessing | 100 | 11,602 | 15,383 (not empty) |
| Type of text units | Small text | Pairs of small texts | Pairs of small texts |
| Similarity marks | 4 levels | 0/1 | 0/1 |
| Source of judgement | Predefined | Three judges | Crowdsourcing |
| Words in corpus | 21,555 | 212,690 | 4,930,050 |
| Unique words in corpus | 2119 | 15,851 | 66,412 |
| Max(words) in input documents | 532 | 34 | 4993 |
| Mean(words) in input documents | 215.55 | 19.43 | 320.49 |
| StDev(words) in input documents | 77.82 | 5.26 | 272.76 |
| Words in cleaned corpus | 20,167 | 202,063 | 4,711,213 |
| Max.(words) in cleaned documents | 490 | 32 | 4715 |
| Unique words in cleaned corpus | 1669 | 12,808 | 53,325 |
| Mean(words) in cleaned documents | 201.67 | 18.46 | 306.26 |
| StDev(words) in cleaned documents | 72.67 | 5.02 | 260.51 |
| Lemmatized words | 2404 | 37,491 | 465,592 |
| Morphed words | 2140 | 39,409 | 528,692 |

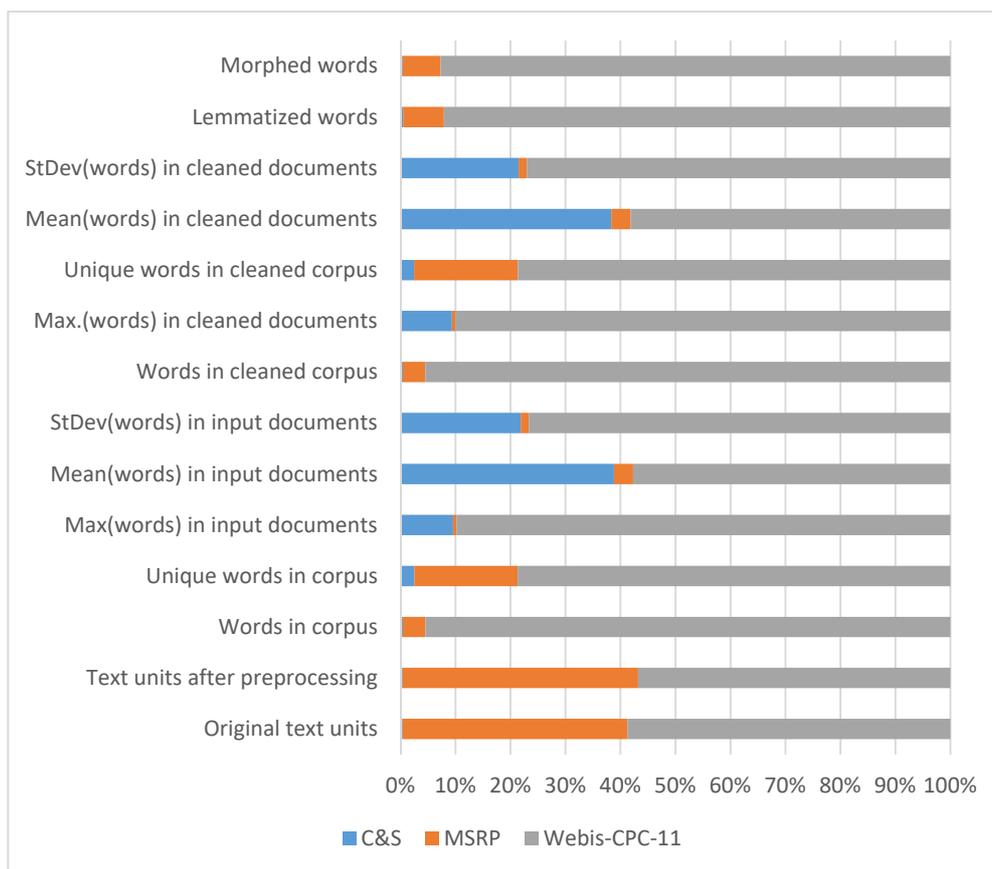

**Figure 2.** Corpora metadata.

Figure 3 shows the word number distributions in the three corpora we used after pre-processing. The document numbers are on the *x*-axis, and the numbers of words contained in the documents are on the *y*-axis. The three corpora are very different in number of text units (100/11602/15718) and in word number (rather uniform in MSRP with maximum of 34 words in one text; there were 532 in C&S; there was an extremely uneven distribution in Webis with maximum of 4993 words in a single text).

This diversity, while seemingly undesirable, has allowed us to come to some somewhat surprising conclusions (presented in Section 5).

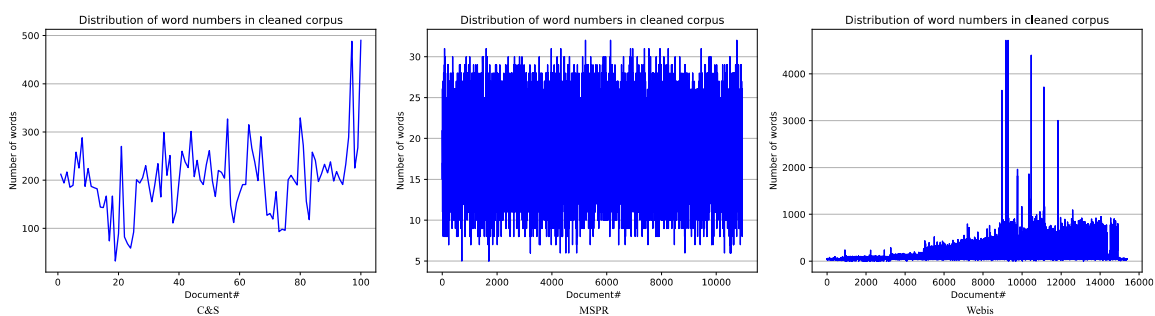

**Figure 3.** Document word number distributions in corpus files.

3.4.3. Text Pre-Processing

Before using various models for word processing, text corpora need to be prepared or pre-processed to enable and to maximize the processing effect.

The texts were pre-processed through quite a number of steps, the major ones of which are presented in the flowchart (Figure 4) and summarized in Table 2 (the options that gave better results are in bold). Some details or steps cannot be presented in the flowchart because that would make it too large and unclear. The steps are defined by the initial parameters written in the plain text configuration file. Actual parameters can be changed from initial ones via the menu.

The text pre-processing process is roughly outlined in the following steps:

1. Determination of empty files and their registration. Empty files are the reason for the differences in the number of documents before and after text pre-processing.
2. Extraction of plain text from documents (parsing). Some files contained problematic text encoding, especially in the Webis corpus. Every symbol out of utf-8 was transcoded into it or deleted.
3. Three variants of number manipulation were tested: deleting them, replacing them with the <NUM> token, and leaving them in the text. Experiments have shown that the best results are achieved by leaving them.
4. Bigrams can affect model results. For bigram formation we used Phrases from genism model's module. Experiments have shown that the results are better without the use of bigrams, so there was no need to implement n-grams.
5. Filtering text with specific part of speech tags (POST) did not produce better results in several POST combinations. For POS filtering, we used the tagger from the nltk package.
6. Stop word processing went in three directions. We tried to (1) delete them, (2) replace them with the token <STOPWORD>, and (3) keep them all. Stop words were left in documents because leaving them gave us better results. This is in contrast to the common and logical practice of text pre-processing, which produces worse results, we suppose because the context of the word is thus preserved. Numbers were left in the text for the same reason.
7. Word lemmatization using WordNet Lemmatizer. Lemmatization is the process of converting a word to its base form—grouping together words with similar meanings but in different inflected forms. While stemming just removes the last few characters, often leading to incorrect meanings and spelling errors, lemmatization correctly identifies the base forms of words. With the lemmatization, we obtained slightly better results than without it.
8. Removing single character words because using them does not improve results, but increases the complexity of models and calculations.

9. Morphing words using Morphy from WordNet. Morphology in WordNet uses two types of processes to try to convert the string passed into one that can be found in the WordNet database. That slightly improved our results.
10. Stemming with three different stemmers (Snowball, Porter, and Porter2), but the stemmers did not contribute to the results, on the contrary.
11. Usage of hypernyms and holonyms from WordNet did not improve results.

**Table 2.** Valuated techniques of text pre-processing.

| Technique | Variants\|Comments |
|---|---|
| Parsing | Necessary |
| Transcoding | UTF-8 |
| Tokenization | Numbers and stop words replaced with token |
| Phrasing (bigrams) | Yes/**No** |
| Part-of-speech tagging | 'NN', 'NNS', 'NNP', 'NNPS', 'VB', 'VBD', 'VBG', 'VBN', 'VBP', 'VBZ', 'JJ', 'JJR', 'JJS', 'RB', 'RBR', 'RBS'/**Not used** |
| Conversion | Delete/replace/**keep** |
| WordNet Lemmatization | **Yes**/No |
| WordNet Morphing | **Yes**/No |
| Stemming | Snowball/Porter/Porter2/**None** |
| WordNet Hyper/Holonym | Hypernym/Holonym/**None** |

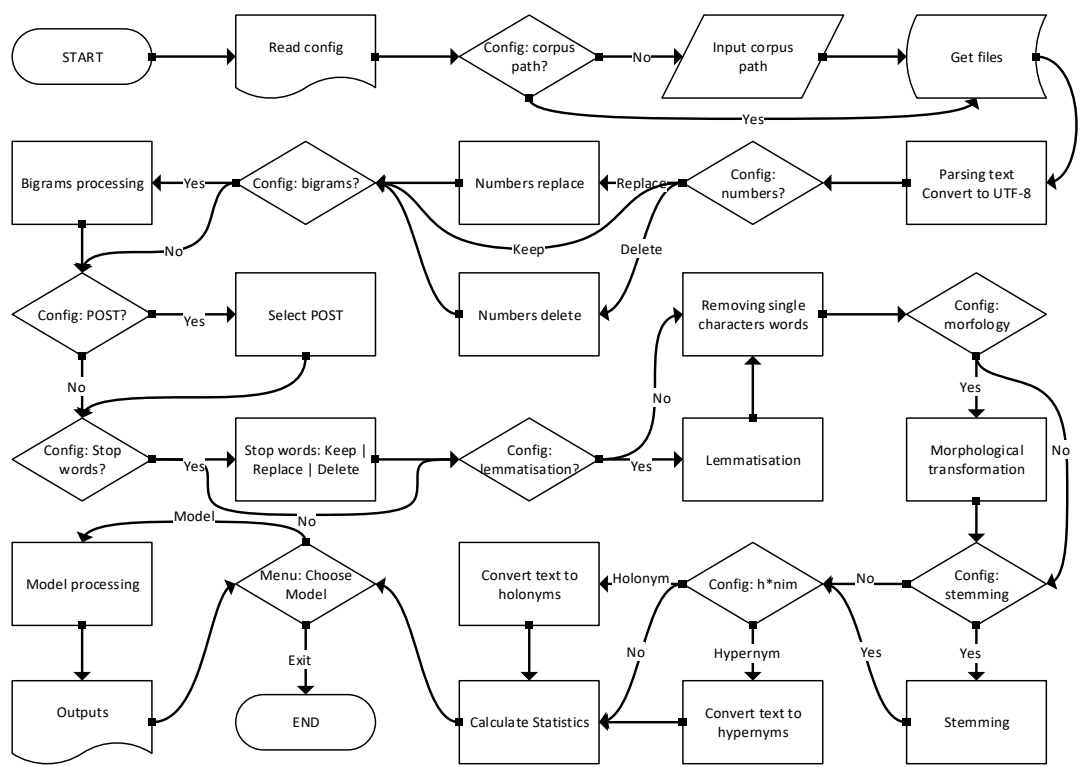

**Figure 4.** Text processing flowchart.

## 4. Results and Discussion

While processing the calculations of eight models over three corpora, it became obvious that the models' results could not be compared while using the same threshold value for all of them, as we intended while planning the experiments. It was necessary to determine the best threshold value for every model and corpus. Threshold values were calculated from the training parts of datasets and evaluated on the testing parts using the 3-cross validation method for Webis; for MSRP we used predefined train and test datasets; and for C&S we used the 5-cross validation method because of the small corpus and the fact that there are five topics in it. The standard measures for different threshold values were calculated, from 0 to 1 in 0.01 steps. Then, the threshold for each individual model was selected according to the best F-measure value from train part of datasets. The summarized data of the evaluations carried out on the test parts of datasets, for results of experiments on eight models and three corpora, are shown below, with figures and numerical values in the tables. For better understanding of the results, we used:

Tabular presentation of calculated thresholds and standard evaluation measures accuracy, precision, recall, and F1.

- Line diagram—comparative representation of standard measures.
- Precision/recall diagram (Appendix A, Figures A1, A3 and A5).
- Receiver operator characteristic curve (Appendix A, Figures A2, A4 and A6).

The best values (bolded in Table 3) can also be visually confirmed in Figures 5–7.

**Table 3.** Cosine similarity evaluation for models at their best thresholds.

| Model | Threshold | | | Accuracy | | | Precision | | | Recall | | | F1 | | |
|---|---|---|---|---|---|---|---|---|---|---|---|---|---|---|---|
| | CS | MS | W | CS | MS | W | CS | MS | W | CS | MS | W | CS | MS | W |
| USE | 0.73 | 0.72 | 0.61 | **0.988** | 0.673 | **0.577** | **0.965** | 0.675 | **0.551** | 0.974 | 0.965 | 0.995 | **0.968** | 0.794 | **0.709** |
| TF-IDF | 0.1 | 0.37 | 0.07 | 0.981 | **0.706** | 0.573 | 0.941 | **0.697** | 0.548 | 0.965 | 0.976 | 0.994 | 0.951 | **0.813** | 0.707 |
| Doc2Vec [2] | 0.7 | 0.39 | 0.48 | 0.908 | 0.655 | 0.569 | 0.862 | 0.655 | 0.546 | 0.634 | **1.0** | 0.994 | 0.726 | 0.792 | 0.705 |
| Glove-D | 0.74 | 0.42 | 0.47 | 0.856 | 0.657 | 0.569 | 0.619 | 0.657 | 0.545 | 0.722 | 0.998 | **0.999** | 0.662 | 0.792 | 0.706 |
| LSI-LSA | 0.42 | 0.54 | 0.29 | 0.899 | 0.679 | 0.572 | 0.973 | 0.673 | 0.548 | 0.497 | 0.99 | 0.995 | 0.654 | 0.802 | 0.707 |
| Word2Vec [1] | 0.63 | 0.83 | 0.9 | 0.721 | 0.69 | 0.572 | 0.412 | 0.687 | 0.548 | **1.0** | 0.966 | 0.996 | 0.582 | 0.803 | 0.707 |
| ELMO | 0.77 | 0.48 | 0.77 | 0.561 | 0.655 | **0.577** | 0.317 | 0.655 | 0.55 | 0.965 | **1.0** | 0.996 | 0.472 | 0.792 | **0.709** |
| FastText | 0.33 | 0.79 | 0.86 | 0.194 | 0.676 | 0.572 | 0.194 | 0.671 | 0.547 | **1.0** | 0.991 | 0.994 | 0.325 | 0.800 | 0.706 |

Legend: [1] Word2Vec-CBoW; [2] Doc2Vec-DBoW.

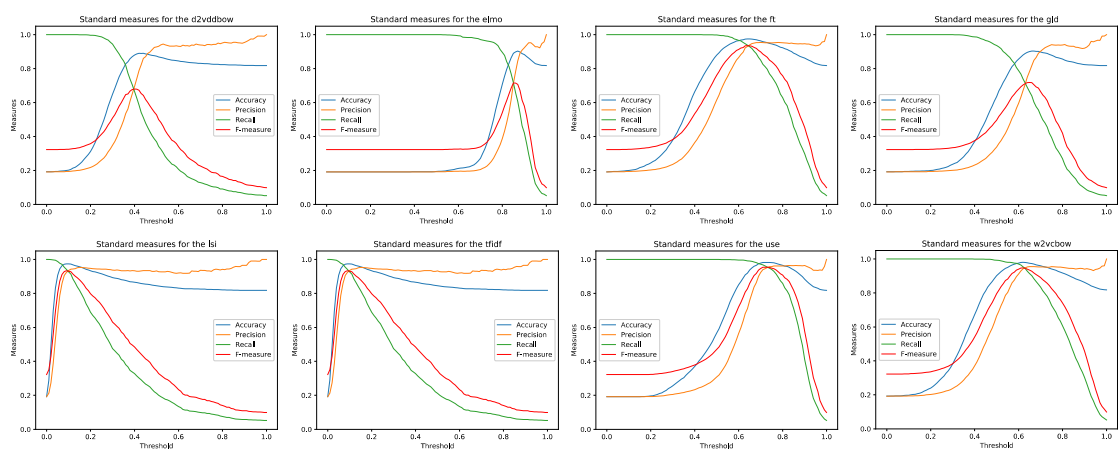

**Figure 5.** Standard evaluation measures for eight models processing C&S corpus.

Figures 5–7 and Table 3 reveal that the best thresholds and standard evaluation measures values from training datasets are quite diverse between different models in the same corpus. By comparing the results of the same model on different corpora, certain unexpected phenomena are observed. Specifically, all models perform worse on the two larger corpora where the top of the F1 curve is almost lost. While the results for the C&S corpus are excellent, with the evaluation curves in line with expectations, the results of all the MSRP and Webis corpora evaluations are almost unchanged from threshold of 0 up to 0.5 or even higher. After the aforementioned threshold, instead of achieving some expected (as seen in Figure 5) growth and maximum, before the final fall in threshold = 1, in classical models (TF-IDF and LSI) the results of evaluation measures fall instead of reaching the maximum

(DL models nevertheless show the maximum of the F1 curve, although barely visible). This might lead us to conclude that the models or some official paraphrase annotations contained in the corpora are invalid. Both possible explanations seemed improbable for two reasons: (1) DL models should produce better results based on a larger set of inputs that train their neural networks; (2) official annotations for corpora used for years by other researchers are expected to be reliable. In search of an explanation, we have done multiple checks on the experiment settings and program code, checks on test corpora (hand-made and surely correctly annotated), and numerous repetitions of experiments with different parameters. We also conducted a manual check of the differences between the ratings produced by the models, a check of the official annotations from the corpus, and a personal assessment of whether or not the text was a paraphrase. We noticed that some examples which were tough ones to judge even for humans, had lexical patterns that were close, but the semantics were different. In MSRP there are also cases where one sentence is the negation of the other one. Lexically, we may have only one difference (the negator: e.g., "not") but sentences pointing in semantic opposite directions. In Appendices B and C, there are some interesting pairs of sentences observed from 25 starting pairs from Webis and MSRP corpora. After presenting the results of our experiments, we are required to present the results obtained by other researchers, which we have introduced in Table 4.

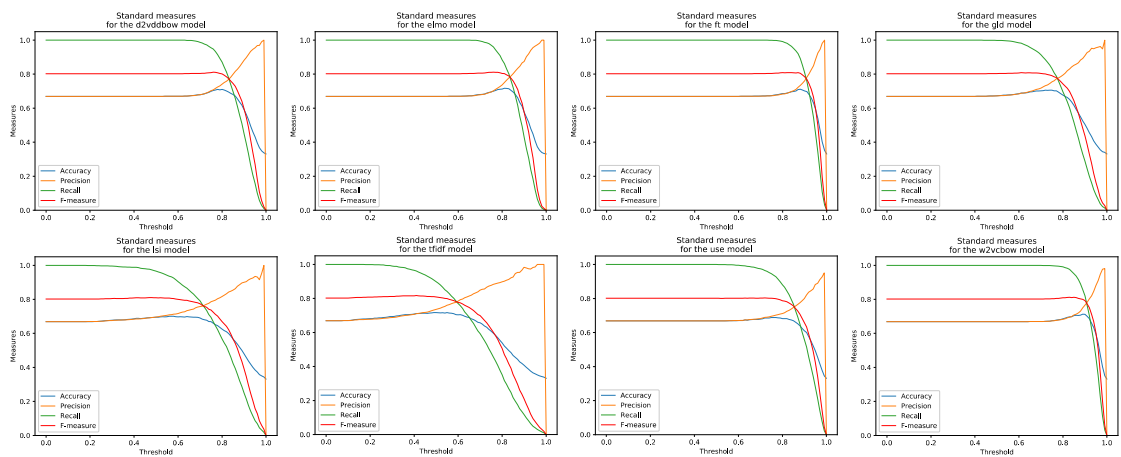

**Figure 6.** Standard evaluation measures for eight models processing MSRP corpus.

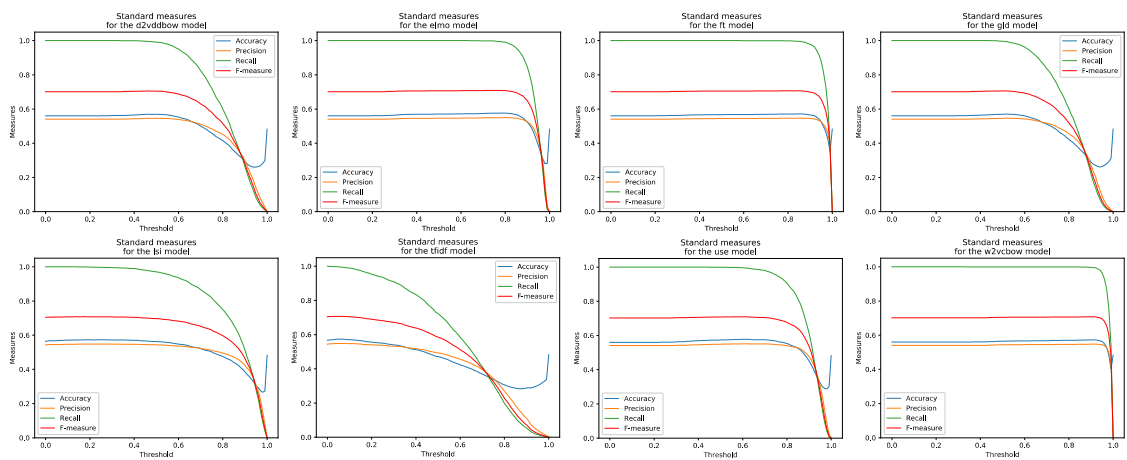

**Figure 7.** Standard evaluation measures for eight models processing Webis corpus.

Table 4. Results of some previous research on unsupervised identification of text similarity.

| ID | Corpus | Base | Approach | Thr | F (%) |
|---|---|---|---|---|---|
| [12] (2005) | MSRP | Words | 6K | 0.5 | 81.2 |
| [13] (2006) | MSRP | Words | 2C, 6K | 0.5 | 81.3 |
| [62] (2007) | MSRP | Sentence, Paragraph | 1C　1K | 0.6 | 81.3 |
| [14] (2008) | MSRP | Words | 6K | 0.8 | 82.4 |
| [63] (2008) | MSRP (balanced) | Words, Phrase, Text | lexicon-syntactic | NS | 68 |
| [17] (2011) | Gigaword (t), MSRP (e) | Words, Phrases | 1C | 0.5 | 83.6 |
| [64] (2011) | Wikipedia (t) | Text, Concepts | 1C, 1K | 0.65-0.95 | 81.4 |
| [19] (2013) | PAN-PC-10 | 5-g Words | 4C, 1K | 0.03 | 68.74 |
| [19] (2013) | PAN-PC-10 | 5-g Words | 6C | NSF | 81.1 |
| [19] (2013) | PAN-PC-10 | Paragraph | 6C | NSF | 99.2 (c), 92.6 (p) |
| [19] (2013) | C&S | 3-g Words | 7C | NSF | 96.2 (c), 97.3 (p) |
| [65] (2014) | MSRP | Sentence | 4C　1K | NSF | 82 |
| [66] (2015) | MSRP, BNC | Words, Sentences | 2C, 1K | NSF | 85.3 |
| [67] (2015) | MSRP | n-gram Words | 1C, 2K | NSF | 84.73 |
| [1] (2017) | MSRP | Words, Sentences | 4C, 3K | NSF | 84.5 |
| [28] (2017) | Webis | Paragraph | 3C | NSF | 86.5 |
| [28] (2017) | C&S | Paragraph | 3C | NSF | 89.7 |
| [1] (2017) | SemEval 2015 Twitter | Words, Sentences | 4C, 3K | NSF | 75.1 |
| [31] (2018) | MSRP | Phrases | 5C | SNLI | 82.3 |
| [35] (2019) | MSRP, BookCorpus | Sentences | 20C | NSF | 83.5 |

Legend: C = corpus-based measure; K = knowledge-based measure; Thr = threshold; NSF = "not specified"; F = F-measure; BNC = British National Corpus; SNLI = Stanford Natural Language Inference; WaCky = "The Web-As-Corpus Kool Yinitiative" (https://wacky.sslmit.unibo.it/doku.php); (c) = "clean"; (p) = "plagiarized"; (t) = "for training"; (e) = "for experiments".

Table 4 gives an overview of some previous research: corpus used, base of metrics (words, sentence, and paragraph), knowledge or corpus based, threshold, and F-measure evaluation. Please note that knowledge-based measures used WordNet as their lexical database source and very few of them used Paraphrase Database. The best results regarding the F-measure are achieved when researchers split the experiment or its evaluation (cannot be deduced from the text) into two parts, one paraphrased and the other not. We see no reason for this separation unless it artificially achieves better results. In real life, this separation does not happen. Should we take these results seriously? This is also why the best results are not bold. Furthermore, it is difficult to explain the fact that many researchers have not documented such important information about the threshold value that they used to define a text as a paraphrase or not. We can assume that they used a threshold that gives the best results.

## 5. Conclusions

This paper presents a systematic study of corpus-based models that are capable of determining the semantic similarity of texts and can therefore be used in the task of paraphrasing detection. More precisely, we experimented with different text representation models by varying text pre-processing scenarios, hyper-parameters, sub-model selection, distance measures, and semantic similarity/paraphrase detection threshold. We selected six deep learning models, Word2Vec, Doc2Vec, GloVe, FastText, ELMO, and USE, and two traditional models, TF-IDF and LSI. We evaluated these models in terms of accuracy, precision, recall, and F1 measure on three publicly available corpora (Microsoft Research Paraphrase Corpus, Clough and Stevenson and Webis Crowd Paraphrase Corpus 2011).

There are many related studies, articles, and studies that are focused on paraphrasing detection, but to the best of our knowledge, the present study is the first attempt to examine and compare the performances of larger sets of the corpus-based models, on three publicly available corpora with the task of paraphrase detection. In summary, two main contributions of our research are: providing an overview of the performances of corpus-based models in the task of paraphrase detection, and reporting the results based on experimenting with numerous variations of approaches.

Using DL models, we were generally able to obtain better results than can be obtained with the TF-IDF and LSI (LSA) models for most of the evaluation measures we used: accuracy, precision, recall and F1. The exceptions are accuracy and F1 results for the MSRP corpus and TF-IDF model, and recall results for the Webis corpus and LSI/LSA model, but these results are strange because of the low best thresholds under which they were achieved. Why is there such a low threshold for these models? Additionally, why are F1 results fairly even for larger corpora, MSRP and Webis? Are such results a consequence of the size of the corpora or because the corpora also contain such pairs of sentences that are lexically very close but semantically different, as we discovered by manually checking pairs? Can the cause be ascertained? Yes, by removing such pairs from the corpora, and then repeating experiments on the cleaned corpora. But it is a very time-consuming multi-month job. It is worth mentioning that we are not the only ones have had such a problem, as similarly reported by Madnani et al. [68] (p. 187).

However, we think that the results demonstrate that DL models can surely be used to detect paraphrasing. Moreover, the results indicate that the USE model slightly outperforms other models, and surprisingly TF-IDF performs much better than expected. Unfortunately, it becomes clear to us that DL models results by itself are still insufficient to fully solve the problem of paraphrase detection, due to the difficult cases that exist, which depends more on the semantic level than lexical patterns. We should find a way to refine them, for example, by adding not only similarity measures but also dissimilarity measures in modeling paraphrasing detection. This would, for example, make one negation, which changes the semantic meaning of a lexically very similar text, making it easier to recognize, since the measure of dissimilarity would be of high value.

The first step in further research is the formation of a corpus of paraphrased sentences. The existing ones will be used as a starting point. All doubtful pairs of sentences will be removed. If less than 10,000 pairs of sentences remain, said refined corpus will be supplemented with additional pairs. They will be obtained by manually analyzing the paraphrased sentences obtained using several online sites offering paraphrasing services or through Google Translator by translating English to language1; then to language2; and finally, back to English, where language1 and 2 are some of the world's languages well supported by Google. In addition to or instead of Google Translator, there are several web sites that offer quality text paraphrasing. Experiments will then be conducted on the corpus thus obtained. The corpus will be available for free downloading at the same time that the results of the newly conducted research will be released, thereby enabling new research in the field of paraphrase detection. The goals of further research should answer the following questions. How can one get the highest quality sentence vectors from the word models obtained by DL models, given that this is still an unresolved research question after years of research by many researchers? Which measures are best for determining the semantic similarity and dissimilarity of the two sentences represented by the sentence vectors, given the largest F-measure and the smaller need for computational resources? The above experiments will be performed using several DL models used so far, but also new ones that are or will be introduced in the meantime, such as BERT.

**Author Contributions:** The authors worked together on all parts of the research. All authors have read and agreed to the published version of the manuscript.

**Funding:** This research was funded by the University of Rijeka grant number uniri-drustv-18-38.

**Conflicts of Interest:** The authors declare no conflict of interest.

## Appendix A. P/R Diagrams and ROC Curves

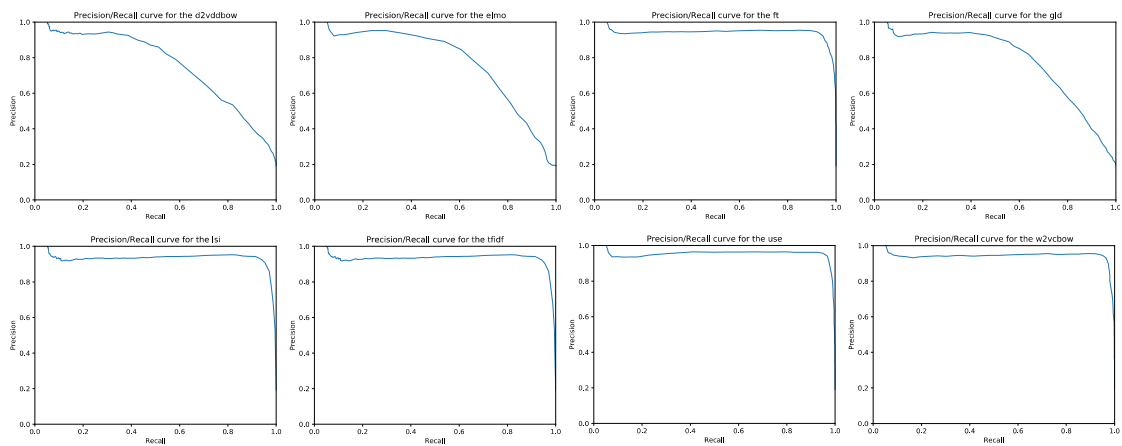

**Figure A1.** Precision/recall curves for eight models processing C&S corpus.

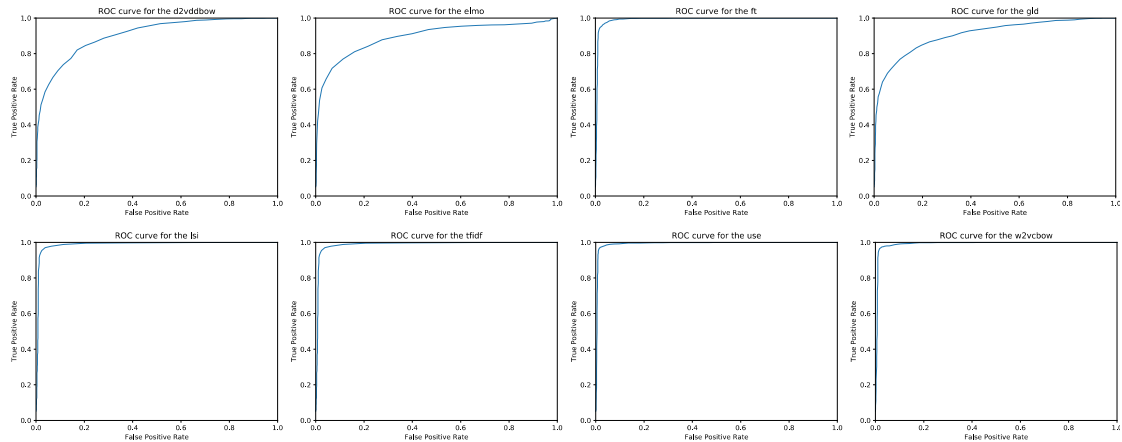

**Figure A2.** Receiver operator characteristic curves for eight models processing C&S corpus.

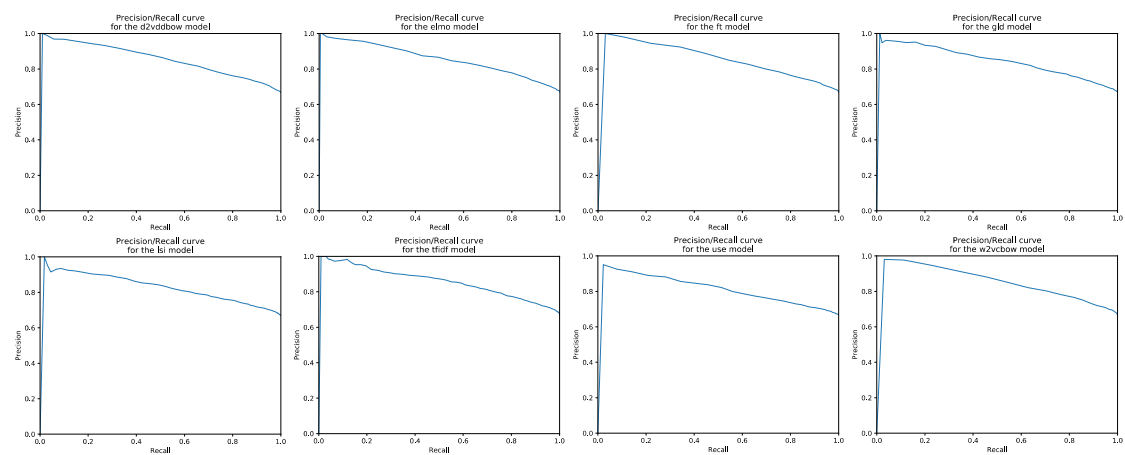

**Figure A3.** Precision/recall curves for eight models processing MSRP corpus.

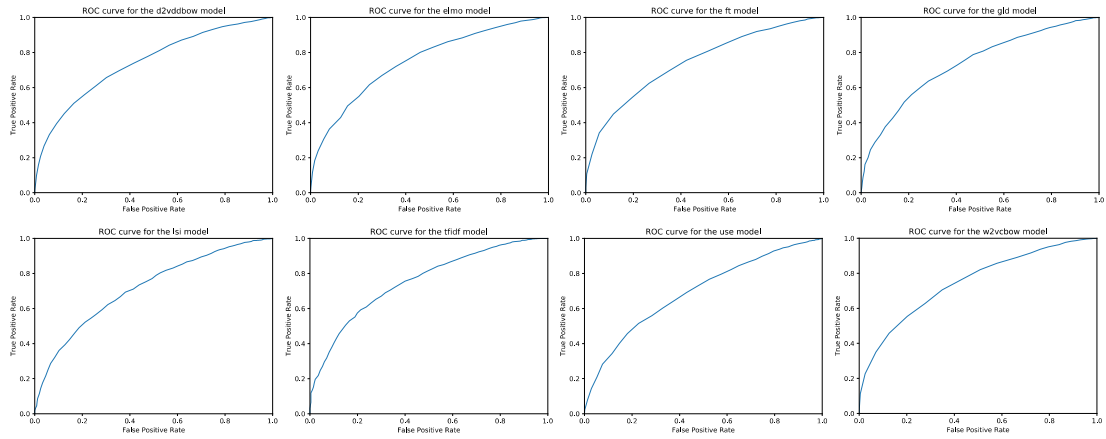

**Figure A4.** Receiver operator characteristic curves for eight models processing MSRP corpus.

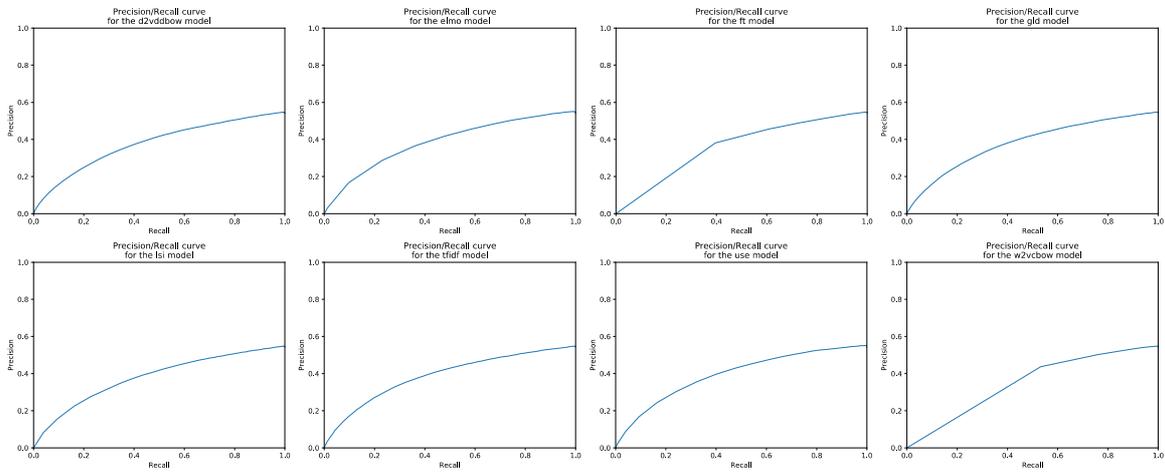

**Figure A5.** Precision/recall curves for eight models processing Webis corpus.

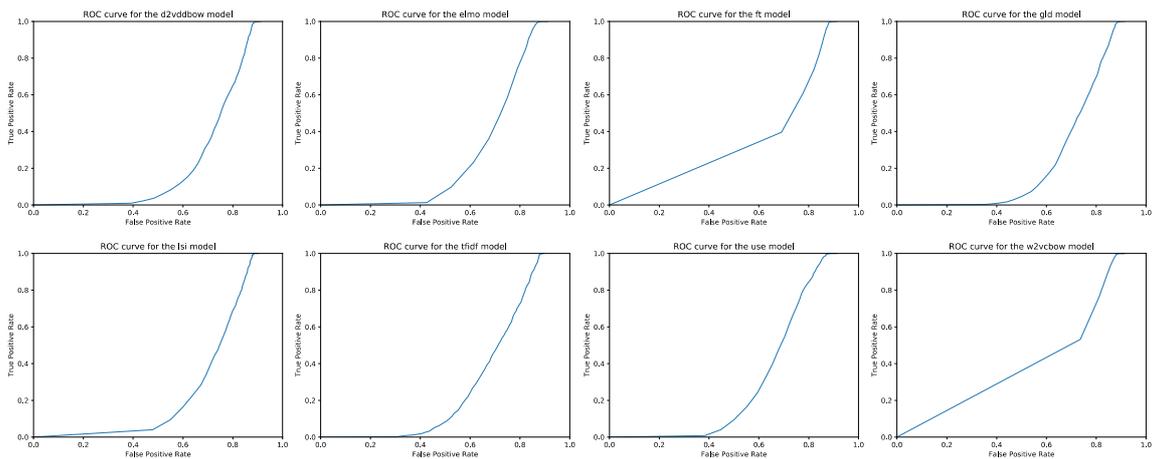

**Figure A6.** Receiver operator characteristic curves for eight models processing Webis corpus.

# Appendix B

Table A1. Some Official Annotations for the Initial 25 Pairs of Webis Corpus.

| # | Original | Paraphrased | O.R. |
|---|---|---|---|
| 5 | "I have heard many accounts of him," said Emily, "all differing from each other: I think, however, that the generality of people rather incline to Mrs. Dalton's opinion than to yours, Lady Margaret." I can easily believe it. | "I have heard many accounts of him," said Emily, "all different from each other: I think, however, that the generality of the people rather inclined to the view of Ms Dalton to yours, Lady Margaret". That I can not believe. | 0 |
| 9 | "Gentle swain, under the king of outlaws," said he, "the unfortunate Gerismond, who having lost his kingdom, crowneth his thoughts with content, accounting it better to govern among poor men in peace, than great men in danger". | "gentle swain, under the king of outlaws," said he, "the fortunate gerismond, who having lost his Kingdom, crowneth his thoughts with the content, accounting it better to govern among poor men in peace, the great men in danger". | 0 |
| 12 | [Greek: DEMOSTHENOUS O PERI TÊS PARAPRESBEIAS LOGOS.] DEMOSTHENES DE FALSA LEGATIONE. By RICHARD SHILLETO, M.A., Trinity College, Cambridge. Second Edition, carefully revised. Cambridge: JOHN DEIGHTON. London: GEORGE BELL. | [Hellene: DEMOSTHENOUS O PERI TÊS PARAPRESBEIAS LOGOS.] Speechmaker DE FALSA LEGATIONE. By RICHARD SHILLETO, M.A., Divine College, Metropolis. Indorse Edition, carefully revised. Metropolis: Evangel DEIGHTON. Author: Martyr Push. | 0 |
| 13 | [Greek: DEMOSTHENOUS O PERI TÊS PARAPRESBEIAS LOGOS.] DEMOSTHENES DE FALSA LEGATIONE. By RICHARD SHILLETO, M.A., Trinity College, Cambridge. Second Edition, carefully revised. Cambridge: JOHN DEIGHTON. London: GEORGE BELL. | [Greek: DEMOSTHENOUS O PERI TES PARAPRESBEIAS LOGOS.] DEMOSTHENES DE FALSA LEGATIONE. By RICHARD SHILLETO, M.A., Trinity College, Cambridge. Second Edition, carefully revised. Cambridge: JOHN DEIGHTON. London: GEORGE BELL. | 0 |
| 21 | "Thus you've heard about the new silk?" said Mrs. Lawton. "To be sure I have," rejoined Sukey. "Everybody's talking about it. Do show it to me, Catharine; that's a dear." The dress was brought forth from its envelope of white linen. | When Sukey saw the unique white dress, she claimed to have no idea what it was made of "But I told you about the new silk yesterday." "You really didn't. I haven't heard a word about it." | 1 |
| 25 | I forgot to mention she added a very minute piece of mace, not enough to make its flavour distinguishable; and that the coffee-pot must be of tin, and uncovered, or it cannot form a thick cream on the surface, which it ought to do. | I forgot to name she supplemental a rattling microscopic fix of official, not sufficiency to modify its sort different; and that the coffee-pot moldiness be of tin, and denuded, or it cannot taxon a jellylike elite on the cover, which it ought to do. | 0 |

Legend: O.R. = official rating.

# Appendix C

Table A2. Some Official Annotations for the Initial 25 Pairs of MSRP Corpus.

| #  | ID1     | Sentence1 | ID2     | Sentence2 | O.R. |
|----|---------|-----------|---------|-----------|------|
| 11 | 222621  | Shares of Genentech, a much larger company with several products on the market, rose more than 2 percent. | 222514  | Shares of Xoma fell 16 percent in early trade, while shares of Genentech, a much larger company with several products on the market, were up 2 percent. | 0 |
| 12 | 3131772 | Legislation making it harder for consumers to erase their debts in bankruptcy court won overwhelming House approval in March. | 3131625 | Legislation making it harder for consumers to erase their debts in bankruptcy court won speedy, House approval in March and was endorsed by the White House. | 0 |
| 13 | 58747   | The Nasdaq composite index increased 10.73, or 0.7 percent, to 1,514.77. | 58516   | The Nasdaq Composite index, full of technology stocks, was lately up around 18 points. | 0 |
| 16 | 142746  | Gyorgy Heizler, head of the local disaster unit, said the coach was carrying 38 passengers. | 142671  | The head of the local disaster unit, Gyorgy Heizler, said the coach driver had failed to heed red stop lights. | 0 |
| 17 | 1286053 | Rudder was most recently senior vice president for the Developer and Platform Evangelism Business. | 1286069 | Senior Vice President Eric Rudder, formerly head of the Developer and Platform Evangelism unit, will lead the new entity. | 0 |
| 18 | 1563874 | As well as the dolphin scheme, the chaos has allowed foreign companies to engage in damaging logging and fishing operations without proper monitoring or export controls. | 1563853 | Internal chaos has allowed foreign companies to set up damaging commercial logging and fishing operations without proper monitoring or export controls. | 0 |
| 19 | 2029631 | Magnarelli said Racicot hated the Iraqi regime and looked forward to using his long years of training in the war. | 2029565 | His wife said he was "100 percent behind George Bush" and looked forward to using his years of training in the war. | 0 |
| 21 | 2044342 | "I think you'll see a lot of job growth in the next two years," he said, adding the growth could replace jobs lost. | 2044457 | "I think you'll see a lot of job growth in the next two years," said Mankiw. | 0 |
| 25 | 1713015 | A BMI of 25 or above is considered overweight; 30 or above is considered obese. | 1712982 | A BMI between 18.5 and 24.9 is considered normal, over 25 is considered overweight and 30 or greater is defined as obese. | 0 |

Legend: O.R. = official rating.